\documentclass[journal]{IEEEtran}

\usepackage{cite}
\usepackage{amsmath,amssymb,amsfonts}
\usepackage{amsthm}
\usepackage{mathtools}
\usepackage{algorithmic}
\usepackage{graphicx}
\usepackage{textcomp}
\usepackage{xcolor}
\usepackage[svgnames]{xcolor}
\usepackage{microtype}
\usepackage[table]{xcolor}
\usepackage{subcaption}
\usepackage{booktabs}
\usepackage{tcolorbox}
\tcbuselibrary{skins}
\usepackage{enumitem}
\usepackage{multirow}
\usepackage{fontawesome5}

\usepackage[hyphens]{url}
\usepackage{hyperref}
\hypersetup{
    colorlinks=true,
    linkcolor=black,
    citecolor=black,
    filecolor=black,
    urlcolor=black,
}

\usepackage[capitalize,noabbrev]{cleveref}

\usepackage[textsize=tiny]{todonotes}

\theoremstyle{plain}

\theoremstyle{definition}

\theoremstyle{remark}

\begin{document}

\title{PseudoAct: Leveraging Pseudocode Synthesis for Flexible Planning and Action Control in Large Language Model Agents}

\author{Yihan~(Logon)~Wen,~Xin~Chen
\thanks{Y. Wen and X. Chen are with the Department of Electrical and Computer Engineering, Texas A\&M University, College Station, TX, USA; Corresponding email: xin\_chen@tamu.edu.}%
}

   

\maketitle


\begin{abstract}
Large language model (LLM) agents typically rely on reactive decision-making paradigms such as ReAct, selecting actions conditioned on growing execution histories. While effective for short tasks, these approaches often lead to redundant tool usage, unstable reasoning, and high token consumption in complex long-horizon tasks involving branching, iteration, or multi-tool coordination. 
To address these limitations, this paper introduces \texttt{PseudoAct}, a novel framework for flexible planning and action control in LLM agents through pseudocode synthesis. 
Leveraging the ability of LLMs to express task-solving strategies as code, \texttt{PseudoAct} synthesizes a structured pseudocode plan that decomposes a task into subtasks and explicitly encodes control flow, including sequencing, conditionals, loops, parallel composition, and combinations of these logic primitives. Actions are then executed by following this global plan, making the decision logic explicit and temporally coherent. This design reduces redundant actions, prevents infinite loops, and avoids uninformative alternative exploration, enabling consistent and efficient long-horizon decision-making. 
Experiments on benchmark datasets show that our method significantly outperforms existing reactive agent approaches, achieving a 20.93\% absolute gain in success rate on FEVER and setting a new state-of-the-art on HotpotQA.



\end{abstract}



\section{Introduction}

Large language model (LLM) agents \cite{bo2024copper,yuanreinforce} have recently demonstrated strong capabilities in solving complex tasks by integrating multi-step reasoning with action execution. Among these, ReAct-style agents~\cite{yao2022react,aksitov2024rest} have emerged as a widely adopted paradigm, combining natural language reasoning with atomic actions to address complex queries. Despite their success, these agents operate in a fundamentally reactive manner, selecting each atomic action based on an ever-growing execution history. This reactive formulation often results in redundant steps, unstable reasoning paths, and high token consumption \cite{xiao2025improving}, particularly in long-horizon tasks involving conditional branching, iteration, or multi-tool coordination.


Several extensions have been proposed to mitigate these limitations. For example, tree-based or deliberative variants, such as Depth-First Search-based Decision Tree (DFSDT)~\cite{qin2023toolllm}, allow agents to explore and backtrack across multiple reasoning paths before committing to an action. While this improves robustness in some scenarios, 
decision-making remains a sequence of local choices, with no explicit representation of future intent, control flow, or task dependencies. As a result, agents often revisit similar states, consume excessive tokens by retracing steps, and struggle to maintain a coherent global structure when execution logic becomes complex. 
More structured approaches have explored explicit planning representations, such as graph-based or hierarchical frameworks~\cite{liu2024codexgraph, erdogan2025planandact,wu2024graph}. However, these frameworks often constrain the agent to follow fixed, pre-defined trajectories and lack the expressiveness needed to represent dynamic workflow control, such as cyclic iteration processes analogous to \emph{for}-loops in code execution. Particularly, these existing methods do not fully leverage one of the core strengths of modern LLMs: the ability to generate structured code for planning and task completion. Unlike unstructured text, code naturally encodes control flow (e.g., \emph{for} loops, \emph{if-else} branching), modularity (e.g., encapsulated function calls), and logical dependencies (e.g., passing output variables to downstream steps).


This motivates a fundamental shift: rather than relying on handcrafted rules or reactive re-planning strategies~\cite{yao2022react, qin2023toolllm}, we propose leveraging the pretrained capabilities of LLMs to generate structured code for planning and action control. Modern LLMs can generate programs with explicit control flow, well-defined variable scopes, and clear input–output dependencies. To harness these capabilities, we adopt pseudocode with control-flow primitives (see Table~\ref{tab:primitives} in Section \ref{sec:overview}) as the planning representation. This allows the LLM to guide reasoning and execution in a structured and flexible manner.



In this paper, we propose a novel framework, \texttt{PseudoAct}, which integrates \emph{Pseudo}code-based planning with guided \emph{Act}ion execution to support flexible decision-making in LLM agents. Specifically, given a user query and action descriptions, the agent first synthesizes a comprehensive pseudocode plan that explicitly defines the subtask structure, control-flow logic, and data dependencies. This plan is created before any actions are executed and serves as a global blueprint for task execution. A pseudocode-guided Control-Flow Executor then follows the plan step by step and uses the predefined execution logic, including termination conditions and state dependencies, to guide a ReAct-style agent in carrying out actions flexibly and efficiently. This formulation establishes a clear separation of concerns: the Planner is responsible for long-horizon structure (e.g., ``repeat this step ten times"), while the Executor manages local execution details (e.g., ``call the API"). By specifying the control flow in advance, the approach avoids the instability of reactive agents that reconstruct their plans at every step.


In summary, our main contributions are threefold:
\begin{itemize}[leftmargin=13pt, noitemsep, topsep=0pt]
    \item [1)] We introduce \texttt{PseudoAct}, a novel framework that reformulates agentic planning and action as pseudocode synthesis. Leveraging the code-generation capabilities of LLMs, this framework generates explicit execution plans that adapt seamlessly to a wide range of task structures, from simple linear sequences to complex control flows involving loops, conditionals, and parallel execution. 

    \vspace{3pt}

    
    \item [2)] Guided by the pseudocode plan, the control-flow Executor orchestrates flexible action sequences while maintaining logical consistency and runtime safety. By managing context and data dependencies, it avoids common failure modes such as infinite loops and redundant exploration, and it significantly reduces token consumption compared to reactive baselines that rely on growing action histories.


    \vspace{3pt}
    
    \item [3)] Extensive experiments on benchmark datasets demonstrate the effectiveness of structured planning with \texttt{PseudoAct}, achieving a 20.93\% absolute improvement in success rate on FEVER \cite{thorne2018fever} and outperforming state-of-the-art methods on HotpotQA \cite{yang2018hotpotqa}. Additional evaluations in practical power grid applications further validate its applicability to complex, long-horizon tasks and underscore \texttt{PseudoAct}'s ability to scale robustly across tasks of varying complexity. 
    

\end{itemize}

\vspace{3pt}

The remainder of this paper is organized as follows. 
Section~\ref{sec:method} introduces the motivations and details of the proposed \texttt{PseudoAct}. Section~\ref{sec:exper} presents evaluation results on two benchmark datasets and practical power grid applications. Conclusions are drawn in Section~\ref{sec:conclusion}. 


\subsection{Related Work}

Recent work on LLM-based agents broadly falls into two categories: 
(1) methods that enhance reasoning and planning during inference, and 
(2) decision-making frameworks that emphasize environmental interaction and long-horizon execution. 
We review both lines below and summarize the key features.

\emph{Language Models for Reasoning.}
Early approaches such as Chain-of-Thought (CoT) \cite{wei2022chain} rely on linear token generation, while more recent methods introduce non-linear reasoning trajectories via structured search or planning. Search-based and graph-based frameworks, such as Think-on-Graph~\cite{sunthink} and SPIRAL~\cite{liu2025spiral},  integrate tree search, graph traversal, or symbolic planning to enable backtracking and multi-hop reasoning. To further improve execution flexibility, action-augmented reasoning methods extend LLMs with external environments. Chain of Code~\cite{li2023chain} executes intermediate reasoning steps via a code emulator but is sensitive to minor syntax errors, whereas ReAct~\cite{yao2022react} uses reasoning to select tool actions, improving robustness and explainability. However, this reactive method relies on fixed action-observation loops, which can lead to redundant steps or non-terminating interactions.
Complementary work on reliability and efficiency, such as conformal language model reasoning~\cite{rubinconformal} and hybrid policy optimization (HiPO)~\cite{deng2025hipo}, improves internal consistency but typically operates independently of explicit execution control.

\emph{Language Models for Decision-Making.}
Beyond reasoning, decision-making agents must interact with environments, maintain states, and execute long-horizon plans.
Recent benchmarks such as TravelPlanner and RefactorBench~\cite{gautamrefactorbench} evaluate agents under realistic constraints, highlighting the need for persistent state management.
From a theoretical perspective, methods like DeLLMa~\cite{liudellma} and analyses of decision geometry~\cite{joshigeometry} study uncertainty and solution-space navigation.
Agentic frameworks further incorporate world or knowledge models to guide planning~\cite{qiao2024agent} or enable self-improvement through trajectory reuse~\cite{sarukkai2025self}.
Despite these advances, most decision-making agents still lack explicit, executable control-flow representations governing iteration, branching, and termination. This limitation motivates the development of our \texttt{PseudoAct} framework.

\begin{figure*}\centering\includegraphics[width=0.85\textwidth]{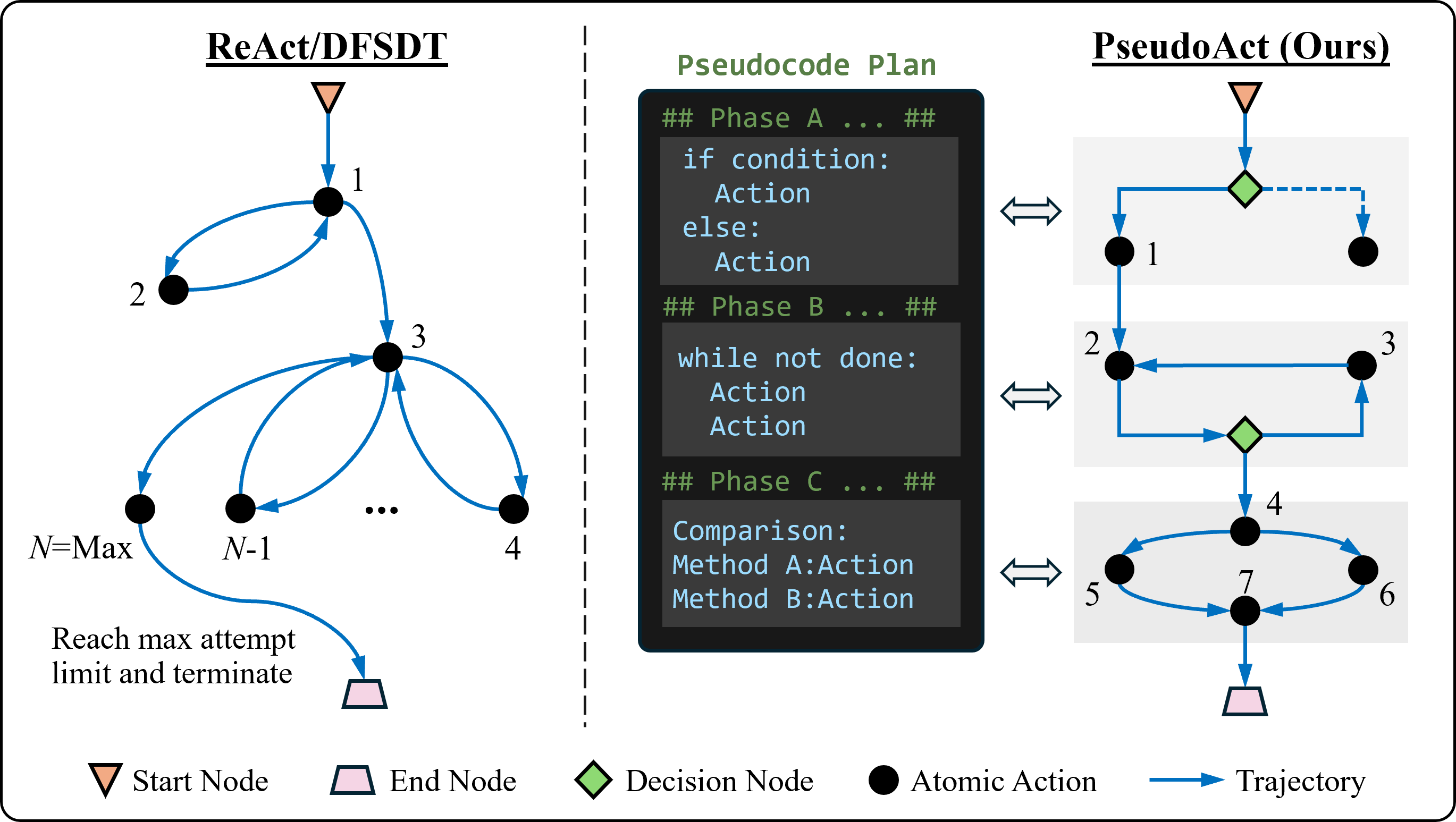}\caption{Comparison between reactive ReAct/DFSDT and our \texttt{PseudoAct} framework. (Reactive agents follow unstructured trial-and-error action sequences and may fail after reaching the maximum attempt limit. In contrast, \texttt{PseudoAct} generates a structured pseudocode plan that decomposes the task into phases with explicit decision logic, guiding action selection and reducing unnecessary exploration).}\label{fig:interface}\end{figure*}

\section{Methodology}
\label{sec:method}

In this section, we first outline the problem formulation and the underlying design rationale, and then introduce the \texttt{PseudoAct} framework.

\subsection{Problem Formulation}

We consider an agentic decision-making framework where a large language model $\mathcal{M}$ is tasked with resolving a natural language query $Q \in \mathcal{Q}$ by interacting with an environment $\mathcal{E}$ via a set of discrete actions $\mathcal{A}$. The environment $\mathcal{E}$ comprises a static knowledge base $\mathcal{K}$, such as a large-scale corpus of documents or technical specifications. In this setting, the environment is partially observable; the agent only accesses specific subsets of $\mathcal{K}$ through the execution of actions (e.g., information retrieval or API calls). The objective is to produce an action sequence $a_{1:T}$ that successfully fulfills the query $Q$, where $T$ denotes the task horizon.

Standard reactive paradigms, such as ReAct, model this process as a sequential mapping $\pi: (Q, h_{<t}) \rightarrow a_t$, where each action $a_t$ at step $t$ is generated based on the current execution history $h_{<t}$. As the task horizon $T$ increases, this approach often suffers from cumulative reasoning errors, redundant action execution, and an inability to maintain global logical consistency across complex control structures.

To address these issues, we introduce \texttt{PseudoAct}, a framework that explicitly decomposes the decision-making process into two distinct phases: \emph{pseudocode plan synthesis} and \emph{logic-guided execution}. First, a planning policy $\pi_{\text{plan}}: Q \rightarrow \mathcal{P}$ maps the input query $Q$ to a structural blueprint $\mathcal{P}$, which encodes global control flow and termination conditions. Next, the execution phase is governed by a conditional policy $\pi_{\text{exec}}: (h_{<t}, \mathcal{P}) \rightarrow a_t$, which selects the action $a_t$ based on the cumulative execution history $h_{<t}$ and the synthesized blueprint $\mathcal{P}$, composed of logic primitives that guide the decision-making process. This formulation reframes the task from making sequential, history-dependent decisions to traversing a synthesized program graph. By leveraging the strong code generation capabilities of large language models to produce a structured prior, \texttt{PseudoAct} effectively reduces redundant trajectories before execution begins.




\subsection{Motivation and Design Rationale}

Reactive reasoning paradigms, such as ReAct, formulate decision-making as a sequence of locally conditioned action selection, where each step is chosen based on an expanding execution history. While effective for short-horizon or linear-sequence tasks, these approaches become inadequate for complex tasks that require conditional branching, iterative refinement, or explicit dependency management. In such cases, the agent lacks a representation of future intent and control flow, forcing it to rediscover global structure implicitly at each step. This often leads to redundant tool invocations, unstable execution under long horizons, and brittle behavior when intermediate outcomes deviate from expectations.

To address these limitations,
we propose leveraging pseudocode as an explicit planning representation. Pseudocode naturally encodes control flow constructs such as conditionals, loops, and termination criteria, as well as data dependencies across subtasks, without committing to low-level implementation details. This representation aligns closely with the pretrained capabilities of modern LLMs, which have been extensively trained on code and program-like structures. By expressing planning as structural pseudocode generation, the LLM agent can reason directly over a global task structure and execution logic, without relying on excessively long planning rules or fixed-length action sequences.

\subsection{Our \texttt{PseudoAct} Framework} \label{sec:overview}

We propose \texttt{PseudoAct}, a new decision-making framework that decomposes agentic behavior into two explicitly separated phases: \emph{Pseudocode Plan Synthesis} and \emph{Guided Execution}. In contrast to reactive reasoning paradigms, such as ReAct, which select an atomic action $a_t$ based solely on a growing execution history $h_{<t}$, \texttt{PseudoAct} formulates decision-making as a programming process. Given a user query $Q$ and a set of actions $\mathcal{A}$, the framework first synthesizes a global Pseudocode Plan $\mathcal{P}$ that specifies the entire control structure required to solve the task.

The synthesized plan $\mathcal{P}$ is expressed using a predefined set of high-level \emph{Logic Primitives} that encode conditional branching, iteration, and structured subtask composition. This plan captures long-horizon reasoning and global dependencies in a structured, human-readable format prior to execution. Once generated, $\mathcal{P}$ is executed by a \emph{Control-Flow Executor} that traverses the plan structure and invokes a local execution agent to complete individual subtasks. 

By decoupling plan synthesis from execution, \texttt{PseudoAct} assigns complementary roles to the two phases: global reasoning, abstraction, and error anticipation are resolved at the planning stage, while precise action decision and environment interaction are handled during execution. This separation of concerns enables more stable long-horizon behavior than purely reactive approaches and leverages the strong code synthesis capabilities of modern LLMs to produce explicit, reusable, and verifiable reasoning structures.

\subsubsection{Phase I. Pseudocode Plan Synthesis and Representation}


The \texttt{PseudoAct} framework formulates decision-making as the synthesis of a structured plan that explicitly encodes global execution logic and state dependencies prior to atomic action execution.

\paragraph{Plan Structure}
Given a natural language query $Q$, the Planner generates a structured plan $\mathcal{P}$ defined as:
\begin{equation}
\mathcal{P} = \left\{ \mathcal{S}, \ \tau, \ \phi_{\text{term}}, \ k_{\max} \right\}
\end{equation}
where $\mathcal{S} = [s_1, \ldots, s_n]$ is an ordered sequence of subtask steps, $\tau \in \{\text{sequential, conditional, iterative, hybrid}\}$ denotes the high-level workflow topology, $\phi_{\text{term}}$ specifies the termination criteria for iterative processes, and $k_{\max}$ provides an upper bound on iterations. These parameters define the global execution constraints applicable to the entire workflow.

Each individual step $s_i \in \mathcal{S}$ is represented as a tuple:
\begin{equation}
s_i = \left( \sigma_i, \ d_i, \ \ell_i, \ \mathcal{I}_i, \ \mathcal{O}_i \right)
\end{equation}
Here, $\sigma_i$ specifies the operational context (e.g., specific servers or action domains); $d_i$ provides a natural language descriptor of the subtask objective; and $\ell_i$ encodes the control-flow logic required to achieve that objective in pseudocode format. The input set $\mathcal{I}_i$ and output set $\mathcal{O}_i$ explicitly define the data interface.

\paragraph{Logic Primitives} To ensure the plan is both expressive and executable, the logic field $\ell_i$ is constructed using a set of seven primitives $\Lambda = \{\lambda_1, \ldots, \lambda_7\}$ that capture fundamental control flow patterns:

\begin{table}[h]
\centering
\small
\caption{Logic primitives $\Lambda$ for pseudocode planning.}
\begin{tabular}{@{}ll@{}}
\toprule
\textbf{Primitive} & \textbf{Semantics} \\
\midrule
\texttt{EXECUTE} & Atomic single-action execution \\
\texttt{IF-ELIF-ELSE} & Conditional branching \\
\texttt{FOR x IN C} & Bounded iteration over collection $C$ \\
\texttt{WHILE $\phi$} & Convergence-driven iteration \\
\texttt{TRY-ON\_FAILURE} & Fault-tolerant execution with fallback \\
\texttt{PARALLEL} & Concurrent independent computations \\
\texttt{DATA-FLOW} & Explicit inter-step data passing \\
\bottomrule
\end{tabular}

\label{tab:primitives}
\end{table}

\begin{figure}[t] 
	\centering
    \includegraphics[width=0.45\textwidth]{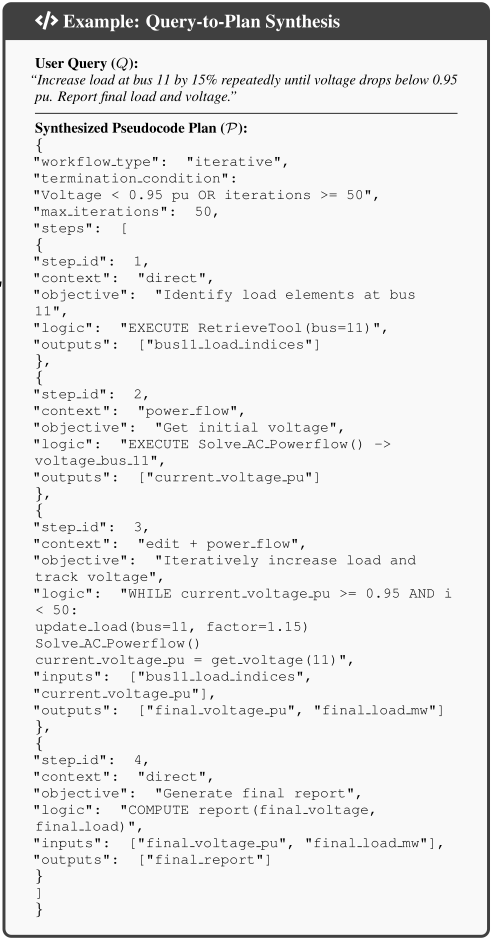}
	\caption{Illustration of a pseudocode plan for resolving a practical multi-step task in the electrical power system domain.
	}
	\label{fig:format}
\end{figure}

Based on these primitives, a long-horizon complex query can be represented as a structured pseudocode plan that integrates control flow and data dependencies in a unified format.
For instance, in the context of electric power systems \cite{grainger1999power}, we consider a multi-step task: incrementally increasing the load demand at a specific location (e.g., bus 11) by 15\%, repeating this process until the system’s voltage magnitude at any point falls below 0.95 per unit (pu). Once this condition is met, the process stops, and the final load level, along with the voltage profile across the power network, is reported.
Figure~\ref{fig:format} illustrates the canonical structure of such a plan, showing how individual steps are composed using the logic primitives to form an executable sequence that supports branching, iteration, and information flow across decision stages.


\paragraph{Structural Properties}

As illustrated in Figure \ref{fig:format},
the structure of $\mathcal{P}$ enforces two critical properties during execution:


\begin{enumerate}
    \item \emph{Guided Control Flow:} The logic $\ell_i$ (e.g., the \texttt{WHILE} loop in Step 3) serves as a structured prompt for the agent. It does not strictly enforce hard-coded tool calls but guides the agent's reasoning, allowing flexibility in tool syntax while rigidly enforcing the high-level termination logic (e.g., voltage $< 0.95$).
    \item \emph{Data Dependency Awareness:} 
    The explicit annotation of inputs $\mathcal{I}_i$ and outputs $\mathcal{O}_i$ induces a directed dependency graph. For example, Step 3 is blocked until Step 1 produces \texttt{bus11\_load\_indices}. This mechanism ensures that the agent is next-step-aware, preventing the hallucination of parameters by guaranteeing that all prerequisites are satisfied before a subtask begins.
\end{enumerate}


\subsubsection{Phase II. Pseudocode-Guided Execution}


The second phase of \texttt{PseudoAct} is managed by a deterministic Control-Flow Executor, which translates the symbolic plan $\mathcal{P}$ into concrete interactions with the environment. In contrast to purely reactive agents that entangle global strategy with local action syntax, our executor adheres strictly to the architectural constraints specified in $\mathcal{P}$, delegating only atomic subtasks to a ReAct-style agent. The key components of this execution process are outlined below.

\paragraph{Execution Mechanism} The executor maintains a global memory state $\mathcal{M}$ to store intermediate outputs and traverses the plan step by step, validating data dependencies before execution. For each step $s_i$, the executor constructs a composite context that combines:
\begin{enumerate}
    \item \emph{Global Constraints:} The workflow topology $\tau$, termination criteria $\phi_{\text{term}}$, and iteration bounds $k_{\max}$.
    \item \emph{Local Context:} The specific step objective $d_i$, logic $\ell_i$, and resolved inputs $\mathcal{I}_i$ retrieved from $\mathcal{M}$.
\end{enumerate}

By inputting these global parameters, the agent retains situational awareness of the broader task structure (e.g., recognizing that it is within a \texttt{WHILE} loop) without being encumbered by the full execution history. The executor then passes this composite prompt to the agent for tool invocation.


\paragraph{Prompt Instantiation} 
To illustrate how these constraints are put into practice, Figure~\ref{fig:prompt_example} shows the concrete context constructed for Step 1 of the power grid task in Figure~\ref{fig:format}. The Workflow Context grounds the agent in the overarching iterative objective, while the Step Context specifies the logic and data requirements for the current subtask.

\begin{figure}[h]
\centering 
\begin{tcolorbox}[colback=white, colframe=black, title=\textbf{Constructed Agent Context (Step 1)}]
\small
\textbf{\texttt{[SYSTEM MESSAGE: Global Constraints]}} \\
\texttt{========= WORKFLOW CONTEXT =========} \\
\textbf{Type:} \texttt{Iterative} \textbf{Termination:} \texttt{Loop stops when voltage < 0.95 pu} \textbf{Max iterations:} \texttt{50} \\
\texttt{=====================================}

\vspace{0.2cm}
\textbf{\texttt{[USER MESSAGE: Local Step Context]}} \\
\texttt{========= TASK DESTINATION ==========} \\
\texttt{Identify all load elements connected to bus 11}

\vspace{0.1cm}
\texttt{========== EXECUTION LOGIC ==========} \\
\texttt{EXECUTE RetrieveTool(table='load', filter='bus': 11) to get indices}

\vspace{0.1cm}
\texttt{=========== DATA INTERFACE ==========}\\ 
\textbf{Input Data:} \texttt{None (Initial Step)}  \\
\textbf{Expected Outputs:} \texttt{['bus11\_load\_indices']}

\vspace{0.1cm}
\texttt{=========== INSTRUCTIONS ============} \\
\texttt{... ...}
\end{tcolorbox}
\caption{An example of the runtime prompt constructed by the Control-Flow Executor for solving the power grid task in Figure \ref{fig:format}. The agent receives both the Global Workflow Context (top) and the Local Step Context (bottom), ensuring local actions are consistent with global intent.}
\label{fig:prompt_example}
\end{figure}

\paragraph{Complexity and Token Efficiency} 


Standard reactive agents repeat the entire interaction history at each step, resulting in a token complexity of $\mathcal{O}(n \cdot L)$ for $n$ steps and a history length of $L$. In contrast, \texttt{PseudoAct} generates the global plan once, incurring a one-time cost of $\mathcal{O}(L_{\text{plan}})$. During execution, each step processes a compact context consisting only of the current step’s requirements and global constraints, where $(L_{\text{step}} + L_{\text{global}}) \ll L$. This yields a total token complexity of $\mathcal{O}(L_{\text{plan}} + n \cdot (L_{\text{step}} + L_{\text{global}}))$, offering a significant reduction in computational overhead.


\paragraph{Termination and Safety Guarantees} Infinite loops are a common failure mode in autonomous AI agents. \texttt{PseudoAct} addresses this issue by elevating termination logic to a structural constraint. The planner explicitly specifies termination conditions $\phi_{\text{term}}$ and iteration limits $k_{\max}$. These conditions are enforced deterministically by the execution engine, ensuring that the system halts even if the underlying agent fails to generate a stop token.

\section{Performance Validation}
\label{sec:exper}





 This section presents numerical tests of \texttt{PseudoAct} on two benchmarks, {FEVER} \cite{thorne2018fever} for fact verification and {HotpotQA} \cite{yang2018hotpotqa} for question answering, in comparison with two baseline methods, ReAct \cite{yao2022react} and DFSDT \cite{qin2023toolllm}. Additionally, \texttt{PseudoAct} is deployed in a real-world industrial application, further demonstrating its effectiveness in practical  settings.





For benchmark tests, {Accuracy} and the {F1} score are used as the primary metrics to assess the effectiveness of the evaluated methods.
Specifically, Accuracy is defined as the proportion of correctly predicted instances to the total number of samples. The F1 score is the harmonic mean of precision and recall, offering a balanced measure of a method’s exactness and completeness, which is computed as:
\begin{align}
	\mathrm{F1} &= \frac{2}{\mathrm{Recall}^{-1} + \mathrm{Precision}^{-1}},\\
        \mathrm{Recall} &= \frac{\mathrm{TP}}{\mathrm{TP} + \mathrm{FN}}, \quad 
	\mathrm{Precision} = \frac{\mathrm{TP}}{\mathrm{TP} + \mathrm{FP}},
\end{align}
where $\mathrm{TP}$, $\mathrm{FP}$, and $\mathrm{FN}$ represent the numbers of true positives, false positives, and false negatives, respectively.







\subsection{Tests on Benchmark FEVER: Fact-Verification Queries}


We evaluate our \texttt{PseudoAct}-based agent on the FEVER (Fact Extraction and VERification) dataset \cite{thorne2018fever}, which is a widely used benchmark for assessing a system’s ability to verify natural-language claims against a large text corpus (Wikipedia) and retrieve supporting evidence for its judgment. The input is a short claim in natural language, e.g., ``\emph{The Eiffel Tower is in Berlin}". Based on the Wikipedia evidence, each claim needs to be assigned one of the three labels: \emph{Supported}, \emph{Refuted}, or \emph{Not Enough Info (NEI)}\footnote{During evaluation, we observed that some samples labeled as NEI are now verifiable using the current version of Wikipedia, likely due to page revisions or annotation inconsistencies, such as the claim ``\emph{Anne Rice was born in New Jersey}". We manually inspected a subset of these cases and calibrated the corresponding labels to support a fair comparison.}. Although individual FEVER claims are short, successful verification requires complex reasoning steps, including identifying the right entities, handling alternative names, checking temporal consistency, and carefully ordering search and lookup operations. Reactive agents need to make these decisions at every step, which often leads to redundant or suboptimal queries. In contrast, \texttt{PseudoAct} encodes these strategies in a pseudocode plan before execution, making tool calls coherent and aligned with the overall verification goal.

For example, we consider the claim: ``\emph{In the End has baseball in it}". Under the FEVER protocol, the agent must decide whether Wikipedia contains explicit evidence that the song “In the End” by Linkin Park includes a baseball-related reference. If no such evidence is found, the correct label is NEI. DFSDT initially retrieves the correct entity with \texttt{search}({\texttt{entity} = \texttt{"In the End"}\}), and a direct keyword probe \texttt{lookup}(\{\texttt{keyword} = \texttt{"baseball"}}) correctly returns \texttt{"No more results"}. At this point, the NEI condition is already met: the relevant entity has been located, and no supporting or refuting sentence is found. However, DFSDT does not treat this as a valid termination signal. Instead, it continues to issue a long sequence of additional \texttt{search} calls with surface-level variations such as \texttt{"In the End lyrics"}, \texttt{"In the End music video"}, and other video- or lyrics-related queries. Most of these calls fail with \texttt{"Could not find ..."}, indicating exploration over non-canonical or non-existent entities in the tool space. Although DFSDT eventually outputs the correct NEI verdict, the trace shows substantial unproductive exploration after the correct evidence state has already been reached. This  highlights a key limitation of DFSDT’s reactive depth-first exploration: it lacks a structural notion of completion once the required evidence condition is satisfied, and thus may continue issuing tool calls in locally unproductive regions of the search space.
In contrast, our \texttt{PseudoAct} method uses explicit pseudocode planning to define both the target evidence form and a termination condition. The plan first grounds the canonical entity (i.e., the song page), then probes for the required evidence slot (e.g., ``baseball" or related sports references). Once the slot is exhausted without finding supporting or refuting evidence, the plan terminates immediately with an NEI verdict. This structured control flow avoids unnecessary entity reformulation and 
enforces clear stopping rules once sufficient evidence has been collected, thereby reducing unnecessary exploration.


The quantitative test performance of \texttt{PseudoAct} is summarized in Table~\ref{tab:hahaha}, in comparison with DFSDT and ReAct. It shows that our method significantly outperforms the ReAct and DFSDT baselines, achieving an accuracy of 88.24\% and an F1 score of 83.35\%. Compared to the strongest baseline (DFSDT), our approach improves accuracy by an absolute margin of 20.93\% and F1 by 19.19\%. These improvements indicate that pseudocode-guided control flow helps stabilize the reasoning process and enhances retrieval precision. The weaker performance of ReAct further underscores the limitations of conditioning action selection solely on execution history, as it frequently becomes trapped in redundant search cycles that our structured planning can avoid.

\begin{table}[t]
\centering
\caption{Quantitative tests and comparison results on the FEVER and HotpotQA benchmarks.} 
\label{tab:main_results}
\normalsize
\begin{tabular}{lccc}
\toprule
& \multicolumn{2}{c}{ {FEVER}} &  {HotpotQA} \\
\cmidrule(lr){2-3} \cmidrule(lr){4-4}
Method & Accuracy (\%) & F1 (\%) & Accuracy (\%) \\
\midrule
ReAct & 60.78 & 62.63 & 46.40 \\
DFSDT & 67.31 & 64.16 & 73.21 \\
\texttt{PseudoAct}  & \textbf{88.24} & \textbf{83.35} & \textbf{82.14} \\
\bottomrule
\end{tabular}
\label{tab:hahaha}
\end{table}

\subsection{Tests on Benchmark HotpotQA: Multi-hop Queries}

HotpotQA \cite{yang2018hotpotqa} is a widely used question-answering benchmark dataset for testing multi-hop reasoning in natural language processing.
Unlike answering a question based on a single sentence or paragraph, each instance in HotpotQA\footnote{In the HotpotQA dataset, we identified a subset of instances with incomplete, ambiguous, or inaccurate ground truth answers. Accordingly, we conducted a manual calibration process to address these issues; see Appendix~\ref{app:calibration} for details.} typically requires multi-hop reasoning and integrating information across multiple documents to derive the correct answer. 
 
In this multi-hop question-answering benchmark, reactive agents such as ReAct and DFSDT often exhibit redundant retrieval operations and unstable execution trajectories as the number of reasoning steps increases. In contrast, our approach explicitly encodes the multi-hop structure and bridge-entity discovery within a pseudocode plan, enabling dependency-aware and structurally constrained execution. 
We take the following bridge-style question as an example to illustrate the distinction: 

\emph{``Seven Brief Lessons on Physics was written by an Italian physicist that has worked in France since what year?''}

This question cannot be answered from a single page; instead, it requires a multi-hop reasoning chain:
\begin{enumerate}
    \item Search the book \emph{Seven Brief Lessons on Physics}.
    \item Identify its author: \emph{Carlo Rovelli} (the bridge entity).
    \item Search the Wikipedia page of \emph{Carlo Rovelli}.
    \item Find the sentence describing his work in France:
    \emph{``Since 2000 he has been a professor \ldots\ in France.''}
    \item Extract the correct answer: \emph{2000}.
\end{enumerate}
Thus, the correct reasoning structure is:
\[
\text{Book} \rightarrow \text{Author (bridge entity)} \rightarrow \text{Career History} \rightarrow \text{Year}.
\]

For this bridge-style question, although DFSDT successfully identifies the bridge entity, Carlo Rovelli, it subsequently enters an unbounded trial-and-error loop. The agent repeatedly issues lookup calls using minor lexical variations related to ``\emph{France}", such as ``\emph{France work}", ``\emph{France timeline}", or numeric-prefix probes. This behavior results in more than 60 tool calls and ultimately triggers a \texttt{GraphRecursionError}. This failure underscores a fundamental limitation of reactive depth-first exploration. Without an explicit plan that specifies both the expected evidence form and the stopping conditions, the agent may continue generating syntactic variations of the same query, leading to recursive expansion and branch explosion rather than meaningful strategy refinement.

In contrast, our \texttt{PseudoAct} agent successfully obtains the correct answer under a predefined plan that constrains the search to a short verifiable two-hop procedure. First, the plan fixes the bridge entity by extracting the author from the book page and navigating directly to the author page, preventing entity drift. Second, it specifies a target evidence format, replacing repeated semantic reformulations with high-precision structural cues commonly used in biographies (e.g., ``since'' followed by a year), which directly match the answer sentence. Finally, bounded exploration with explicit stopping criteria ensures efficient evidence retrieval and avoids the uncontrolled recursion observed in DFSDT. As a result, predefined planning improves robustness by reducing search drift, converting open-ended guessing into pattern-guided retrieval, and preventing tool-call explosion through deterministic termination.

As shown in Table~\ref{tab:hahaha}, our method achieves an accuracy of 82.14\%, significantly outperforming both the standard and LangChain-based implementations of ReAct and DFSDT. While DFSDT improves upon purely reactive baselines by exploring alternative reasoning paths, it remains susceptible to computational inefficiencies on semantically unproductive branches. In contrast, the predefined plan and global coordination in our method reduce execution overhead and prevent infinite loops. 

\subsection{Tests on Practical Power Grid Applications}


To evaluate the generalization and performance of \texttt{PseudoAct} in complex engineering systems, we extend our experiments to electric power grid operations \cite{chen2022reinforcement,chen2025x}. Unlike text-based benchmarks, grid management requires strict compliance with physical constraints, such as voltage limits and thermal ratings, and demands precise, safety-critical decisions. In this subsection, we construct five representative scenarios that reflect core tasks in real-world grid analysis and control. These scenarios systematically evaluate key reasoning capabilities, including iterative feedback control, error handling, cross-network context switching, action validity verification, and sequential state-dependent decision making. Collectively, they provide a structured assessment of the agent’s ability to plan, execute, and adapt in environments where errors can propagate across system states and safety margins are tight.

\paragraph{Scenario 1: Iterative Voltage Collapse}

\begin{figure}\centering\includegraphics[width=0.5\textwidth]{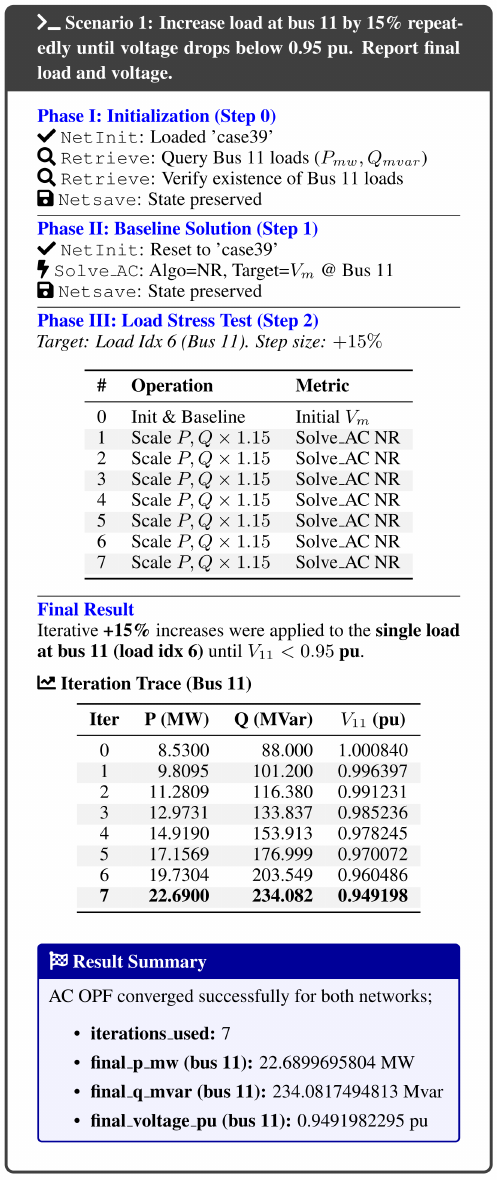}\caption{Overview of Scenario 1. The iteration trace reports the evolution of load and voltage at Bus 11, terminating after seven iterations when the voltage constraint is violated, with the final panel summarizing the converged solution.}\label{fig:interface1}\end{figure}

As illustrated in Figure~\ref{fig:interface1}, the \texttt{PseudoAct} framework adopts a three-stage design to strategically address the voltage collapse task. Phase I (Initialization) begins with a verification step, requiring the agent to confirm the existence of ``Bus 11 loads" before proceeding. This ensures that the target parameters are valid and available for subsequent operations.

\begin{figure}\centering\includegraphics[width=0.46\textwidth]{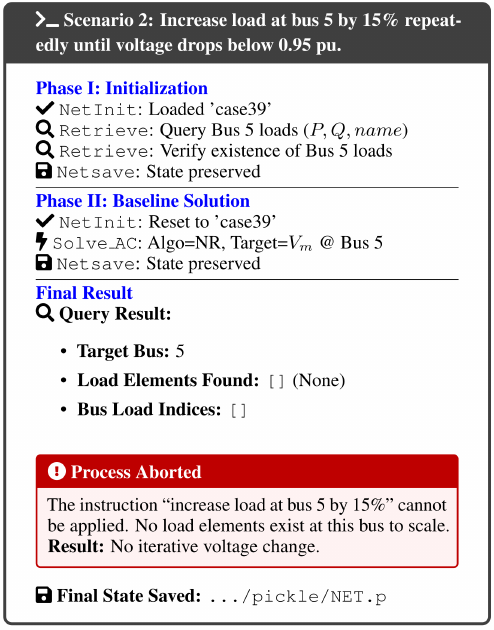}\caption{Scenario 2, in which an attempted 15\% load increase at Bus 5 is aborted because no load elements exist at the target bus, resulting in no voltage variation.}\label{fig:interface2}\end{figure}

\begin{figure}\centering\includegraphics[width=0.46\textwidth]{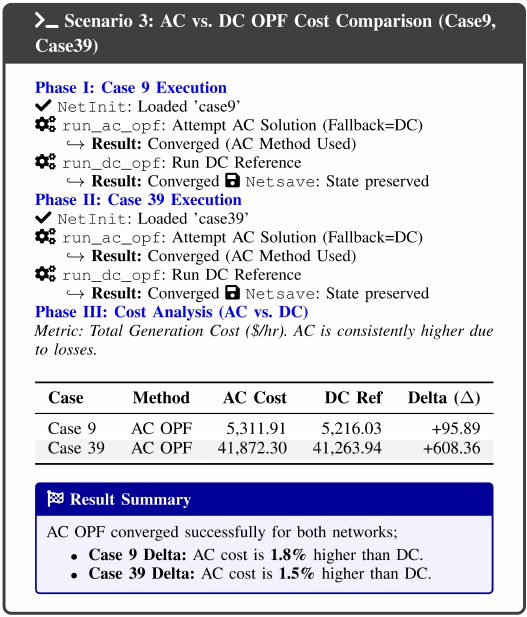}\caption{Comparison of AC and DC optimal power flow (OPF) costs on Case 9 and Case 39, showing successful convergence for both methods and consistently higher AC generation costs due to network losses.}\label{fig:interface3}\end{figure}

\begin{figure}\centering\includegraphics[width=0.46\textwidth]{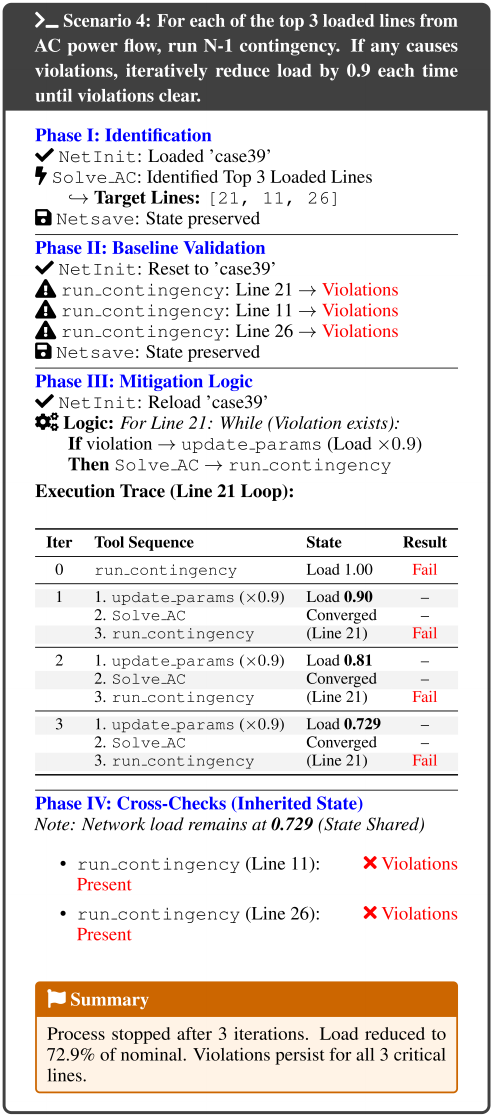}\caption{Scenario 4: Sequential load-shedding mitigation with persistent shared state. The agent identifies the top three overloaded lines, performs iterative load reduction for Line 21 under N-1 contingency, and preserves the updated load state for subsequent evaluations of Lines 11 and 26.}\label{fig:interface4}\end{figure}

\begin{figure}\centering\includegraphics[width=0.5\textwidth]{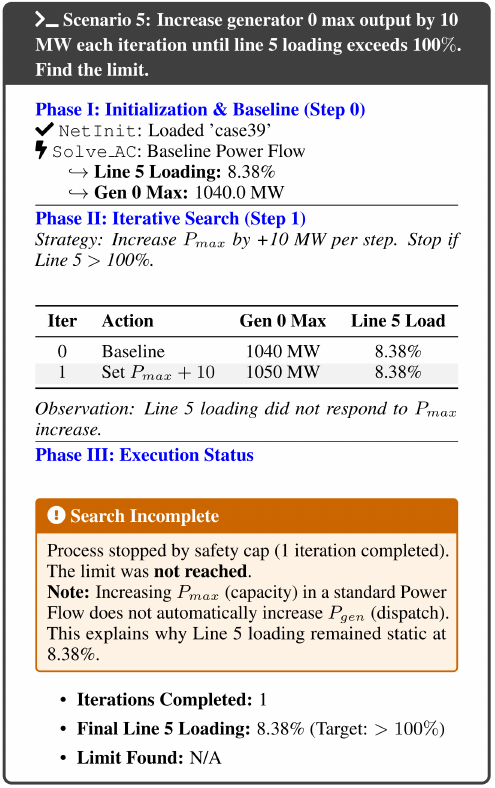}\caption{Scenario 5, illustrating a repeated attempt to increase Generator 0 capacity to induce line overload. The experiment confirms that increasing generation capacity alone does not affect line loading without a corresponding change in dispatch, resulting in no limit being found.}\label{fig:interface5}\end{figure}


\paragraph{Scenario 2: Handling Non-Existent Targets}
As shown in Figure~\ref{fig:interface2}, the proposed \texttt{PseudoAct} framework detects the absence of load elements at Bus 5 during the initialization phase and immediately triggers an abort condition. This prevents invalid modifications and demonstrates robust error-handling capability.

\paragraph{Scenario 3: Comparative Analysis and Context Switching}
We evaluate AC versus DC optimal power flow (OPF) cost comparisons on Case 9 and Case 39 in Figure~\ref{fig:interface3}, under strict state isolation. \texttt{PseudoAct} correctly manages network execution and context switching, ensuring physically consistent results. The AC solutions exceed the corresponding DC costs by 1.8\% and 1.5\%, respectively.

\paragraph{Scenario 4: Sequential Mitigation and State Persistence}
The Figure~\ref{fig:interface5} evaluates the \texttt{PseudoAct}-based agent’s ability to manage nested control flows and mutable system states. The task involved identifying multiple critical components (Lines 21, 11, and 26) and applying an iterative mitigation strategy, specifically load shedding, to each. The execution log highlights two key behaviors. First, the Control-Flow Executor successfully synthesized and executed a nested feedback loop for Line 21, performing three consecutive load reductions ($1.0 \to 0.729$) in an effort to resolve the contingency. Second, the framework demonstrates state persistence: load modifications made during the Line 21 mitigation phase were correctly retained in the shared environment, altering the initial conditions for the subsequent evaluations of Lines 11 and 26.

\paragraph{Scenario 5: Detection of Control Invariance}
As illustrated in Figure~\ref{fig:interface4}, increasing the capacity of Generator 0 had no effect on the loading of Line 5 under AC power flow, which remained constant at 8.38\% without OPF. \texttt{PseudoAct}'s executor detected this control invariance and safely halted execution.






\section{Conclusion}
\label{sec:conclusion}

This paper introduces \texttt{PseudoAct}, a novel framework that redefines agentic decision-making by replacing reactive action selection with structured pseudocode synthesis. By leveraging the code-generation capabilities of large language models, \texttt{PseudoAct} produces global execution blueprints that explicitly encode control flow, data dependencies, and termination conditions. This structured formulation enforces a clear separation between high-level planning and low-level action execution, addressing the reasoning instability and high inference costs inherent to history-dependent, reactive approaches. Empirical evaluations on the FEVER and HotpotQA benchmarks, along with real-world applications in power grid management, show that \texttt{PseudoAct} achieves superior success rates and scalability, highlighted by substantial improvements in accuracy on FEVER and HotpotQA.



\appendix

\subsection{Calibration of Benchmark Dataset HotpotQA} \label{app:calibration}

\begin{figure}[t]
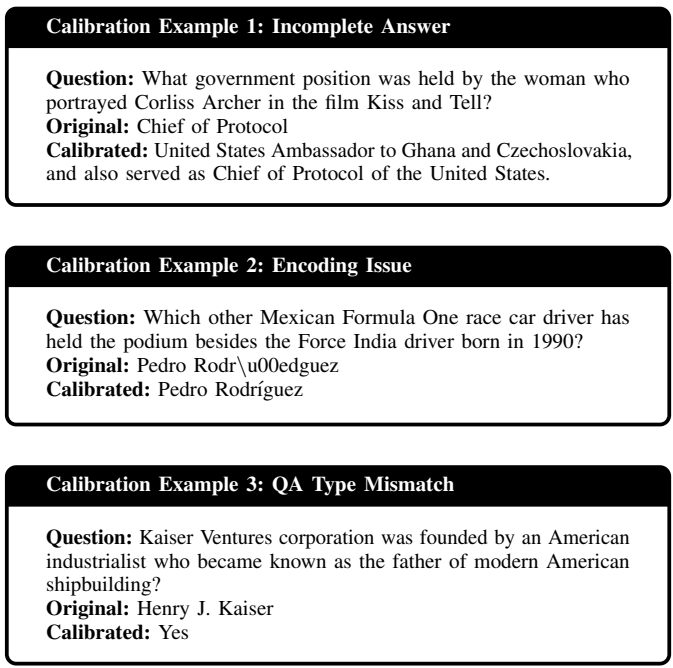

\centering
\footnotesize

\begin{tcolorbox}[colback=white, colframe=black, title=\textbf{Calibration Example 1: Incomplete Answer}]
\textbf{Question:} What government position was held by the woman who portrayed Corliss Archer in the film Kiss and Tell? \\
\textbf{Original:} Chief of Protocol \\
\textbf{Calibrated:} United States Ambassador to Ghana and Czechoslovakia, and also served as Chief of Protocol of the United States.
\end{tcolorbox}

\vspace{0.2cm}

\begin{tcolorbox}[colback=white, colframe=black, title=\textbf{Calibration Example 2: Encoding Issue}]
\textbf{Question:} Which other Mexican Formula One race car driver has held the podium besides the Force India driver born in 1990? \\
\textbf{Original:} Pedro Rodr\textbackslash u00edguez \\
\textbf{Calibrated:} Pedro Rodríguez
\end{tcolorbox}

\vspace{0.2cm}

\begin{tcolorbox}[colback=white, colframe=black, title=\textbf{Calibration Example 3: QA Type Mismatch}]
\textbf{Question:} Kaiser Ventures corporation was founded by an American industrialist who became known as the father of modern American shipbuilding? \\
\textbf{Original:} Henry J. Kaiser \\
\textbf{Calibrated:} Yes
\end{tcolorbox}

\caption{Examples of calibration applied to the HotpotQA dataset, including incomplete answer spans, encoding inconsistencies, and question-answer type mismatches.}
\label{fig:hotpot_calibration}
\end{figure}

During the evaluation process, we identified a subset of instances in the HotpotQA benchmark dataset that contain incomplete, ambiguous, or inconsistent ground truth annotations. To ensure reliable assessment, we conducted a manual calibration process for a number of instances to correct these issues and standardize answer types. Representative examples are provided in Figure~\ref{fig:hotpot_calibration}. 
For instance, in Calibration Example 3, the question ``\emph{Kaiser Ventures corporation was founded by an American industrialist who became known as the father of modern American shipbuilding?}" is syntactically a \emph{yes} or \emph{no} query. However, the original annotation provides an entity answer, ``\emph{Henry J. Kaiser}", resulting in a mismatch between question type and answer format. From a linguistic perspective, the interrogative structure requires a boolean judgment rather than an entity name. We therefore calibrated the ground truth answer to ``\emph{yes}". 
To ensure that the agent demonstrates underlying reasoning rather than simple label prediction, we further require a supporting explanation. For example, the agent produces the reasoning:
``\emph{Henry J. Kaiser was an American industrialist who founded Kaiser Steel and became known for his shipbuilding projects, particularly during World War II. He established the Kaiser Shipyards, which built Liberty ships, and is recognized for his contributions to modern American shipbuilding}". This explanation confirms both factual accuracy and semantic understanding. In this way, the calibration improves evaluation reliability by eliminating answer type inconsistencies and ensuring that correctness reflects genuine reasoning rather than annotation artifacts.

\nocite{langley00}

\bibliographystyle{IEEEtran}
\bibliography{example_paper} 




\end{document}